\documentclass[10pt,twocolumn,letterpaper]{article}

\usepackage{cvpr}              
\usepackage{graphicx}
\usepackage{amsmath}
\usepackage{amssymb}
\usepackage{booktabs}
\usepackage{pifont}
\usepackage{verbatim}
\usepackage{algpseudocode}
\usepackage{algorithm}
\usepackage{algorithmicx}
\usepackage{dsfont}

\usepackage{caption}
\usepackage{amsfonts} 

\definecolor{cvprblue}{rgb}{0.21,0.49,0.74}
\usepackage[pagebackref,breaklinks,colorlinks,allcolors=cvprblue]{hyperref}

\def\paperID{4394} 
\def\confName{CVPR}
\def\confYear{2025}

\title{MetaWriter: Personalized Handwritten Text Recognition Using Meta-Learned Prompt Tuning}

\author{Wenhao Gu, Li Gu, Ching Yee Suen, Yang Wang\\
Department of Computer Science and Software Engineering, Concordia University\\
{\tt\small \{wenhao.gu,li.gu\}@mail.concordia.ca, \{chingyee.suen, yang.wang\}@concordia.ca}
}

\begin{document}
\maketitle

\begin{abstract}

Recent advancements in handwritten text recognition (HTR) have enabled the effective conversion of handwritten text to digital formats. However, achieving robust recognition across diverse writing styles remains challenging. Traditional HTR methods lack writer-specific personalization at test time due to limitations in model architecture and training strategies. Existing attempts to bridge this gap, through gradient-based meta-learning, still require labeled examples and suffer from parameter-inefficient fine-tuning, leading to substantial computational and memory overhead. To overcome these challenges, we propose an efficient framework that formulates personalization as prompt tuning, incorporating an auxiliary image reconstruction task with a self-supervised loss to guide prompt adaptation with unlabeled test-time examples. To ensure self-supervised loss effectively minimizes text recognition error, we leverage meta-learning to learn the optimal initialization of the prompts. As a result, our method allows the model to efficiently capture unique writing styles by updating less than 1\% of its parameters and eliminating the need for time-intensive annotation processes. We validate our approach on the RIMES and IAM Handwriting Database benchmarks, where it consistently outperforms previous state-of-the-art methods while using 20x fewer parameters. We believe this represents a significant advancement in personalized handwritten text recognition, paving the way for more reliable and practical deployment in resource-constrained scenarios. Visit our project page \href{https://chrisgwe.github.io/MetaWriter_PorjectPage/}{here}.

\end{abstract}
\vspace{-13pt}
\section{Introduction}

\begin{figure}
    \centering
    \includegraphics[width=0.95\linewidth]{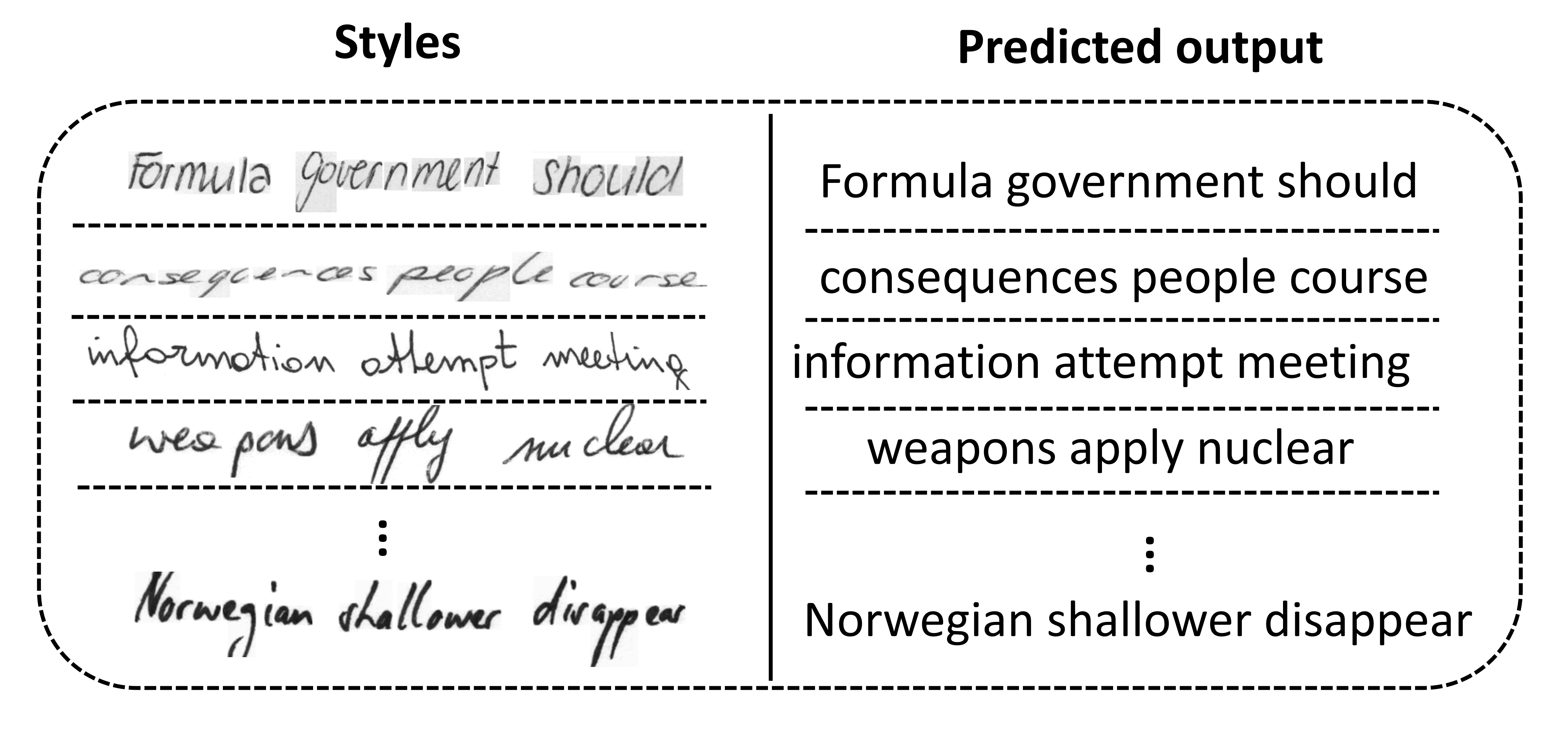}
    \vspace{-10pt}
    \caption{\textbf{Illustration of the challenge of multi-writer handwriting recognition in the IAM dataset.} The left panel shows examples of handwritten text from different writers, highlighting variations in letter shapes, spacing, and stroke patterns. The right panel presents the model’s predictions for each example, demonstrating the difficulty in accurately recognizing diverse handwriting styles. }
    \label{problem_define}
\end{figure}
\vspace{-5pt}

Handwritten Text Recognition (HTR) is the process of converting handwritten text into a machine-readable digital format. As a critical task in computer vision ~\cite{litman2020scatter,luo2020learn,bhunia2021metahtr}, HTR enables the extraction and digitization of handwritten information from diverse sources, including historical manuscripts, forms, whiteboards, memos, etc. The free-flowing nature of handwriting presents inherent challenges, due to irregular character shapes, unstructured layouts, and noise artifacts. In addition, handwritten text exhibits considerable variations in individual styles, influenced by factors such as age, education, linguistic background and ethnicity~\cite{kang2020ganwriting, davis2020text, wang2022fast, bhunia2021metahtr}. For example, some individuals may prefer a pronounced slant, while others favor broader strokes or wider spacing between letters and words, as shown in Figure \ref{problem_define}. However, most recent approaches overlook those common yet complex scenarios, lacking tailored model architectures or training strategies~\cite{coquenet2023dan,fujitake2024dtrocr,coquenet2022end,coquenet2023faster, litman2020scatter,li2019show}. They assume that most variations in handwriting styles are already captured in the training set, leading to poor generalization to unseen styles at test time.


To address this challenge, the work in \cite{bhunia2021metahtr} introduces a test-time adaptation paradigm designed to personalize the model to a specific writer’s style using a few labeled examples collected at test time. For instance, in an iPad application, the system prompts the user to provide a few handwritten words paired with corresponding digital text. By pressing the ``Adapt to My Style'' button, the system automatically adjusts to the user’s handwriting style. Specifically, the proposed method employs a meta-learning algorithm \cite{finn2017modelagnostic} to train a write-adaptive HTR network that aims to adapt the model parameters via a single gradient update step. However, the work in \cite{bhunia2021metahtr} presents two primary drawbacks in real-world deployment scenarios. First, the manual annotation of writer-specific examples for supervised adaptation during deployment is time-consuming and user-unfriendly ~\cite{he2023annollm}. Second, tuning all model parameters imposes substantial computational and memory demands, making test-time adaptation on resource-constrained devices infeasible \cite{cai2020tinytl, lin2022device}.



In this paper, we propose an efficient framework that aims to personalize HTR models for an individual writing style using a few \emph{unlabeled} examples collected at test-time. As a parameter-efficient fine-tuning approach, visual prompt tuning \cite{jia2022visual} enables the model to adapt to downstream tasks by optimizing only the learnable prompt vectors added to the input space, while keeping the rest of the model frozen. Inspired by this paradigm, we incorporate prompt tuning with the assumption that style-specific knowledge can be effectively encoded within these prompt vectors from examples collected at test-time. Therefore, updating the prompts—less than 1\% of the model parameters—enables the model to efficiently capture an unique writing style without incurring substantial computational and memory overhead. To further eliminate the need for labeled examples during deployment, we employ image reconstruction as an auxiliary task using Masked Autoencoder (MAE) \cite{he2022masked} and incorporate the self-supervised objective function to guide the prompt adaptation. This technique significantly reduces reliance on costly, time-intensive annotation processes and enhances adaptability to diverse writer styles in real-world applications. However, the gradient of the prompt vectors with respect to the self-supervised loss for image reconstruction may not align well with the autoregressive loss for text token prediction, potentially introducing instability in the optimization process. To reconcile these losses, we employ a meta-learning algorithm to learn an optimal initialization of the prompts across all writers in the training set, enabling prompt tuning to minimize both the self-supervised and text prediction losses simultaneously at test-time. 

\begin{figure}
    \centering
    \includegraphics[width=0.95\linewidth]{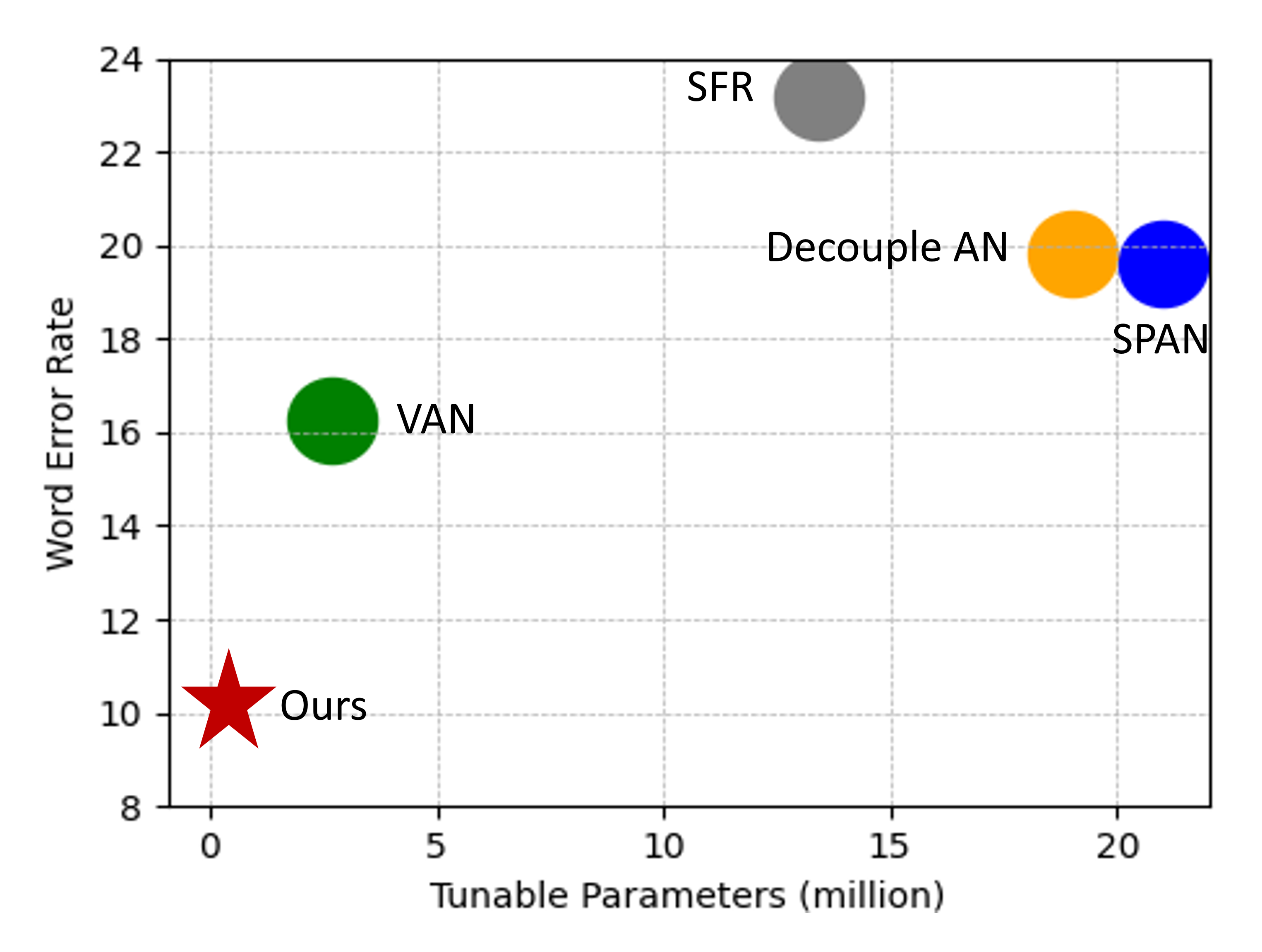}
    \vspace{-16pt}
    \caption{\textbf{Illustration of performance comparison between our method and other state-of-the-art methods.}  We report the word error rate (WER) and the number of tunable parameters (both metrics are better when lower). Our method achieves the lowest WER on the IAM dataset, requiring only a negligible number of tunable parameters for personalization.}
    \label{performance}
\end{figure}



Our work makes three key contributions to advancing personalized handwritten text recognition. First, we introduce an efficient framework that formulates writer-specific personalization as prompt tuning, incorporating an auxiliary image reconstruction task with a self-supervised loss to guide adaptation with unlabeled test-time examples. Second, we leverage meta-learning to optimize prompt initialization, ensuring that prompt tuning at test time minimizes both self-supervised and text recognition errors. Third, our approach outperforms state-of-the-art results on the RIMES and IAM Handwriting database benchmarks and achieves fast adaptation with 20x fewer parameters, paving the way for practical deployment in resource-constrained scenarios. Figure \ref{performance} illustrates the superior performance of our method.

\vspace{-15pt}

\section{Related Works}
We review several lines of research related to our work, including handwritten text recognition, visual prompts, meta-learning, and masked autoencoder.

\noindent{\bf Handwritten Text Recognition}. Recent advancements in Handwritten Text Recognition (HTR) have been shaped by deep learning techniques, particularly convolutional neural networks (CNNs), which enable efficient feature extraction from raw handwritten images \cite{ahlawat2020improved, Smith2021}. The Connectionist Temporal Classification (CTC) layer \cite{graves2006connectionist} enabled end-to-end sequence learning in HTR, later giving way to attention-based decoding that integrated language modeling, character detection, and recognition \cite{lee2016recursive,shi2018aster}. This approach, suited for handling irregular text, has become the state-of-the-art paradigm for both scene text and handwriting recognition \cite{bhunia2019handwriting,luo2020learn,wang2020decoupled,zhang2019sequence}. 
Transformer models with self-attention have advanced handwritten text transcription, effectively handling cursive and unconstrained handwriting without pre-segmentation, aided by large-scale datasets \cite{Anderson2022, Williams2023}.

\noindent{\bf Visual Prompt}. Visual Prompt Tuning provides a parameter-efficient way to fine-tune models by inserting learnable prompt tokens into the visual input, such as image pixels~\cite{bahng2022exploring} or image patch features~\cite{wang2022learning,jia2022visual,huang2023diversity}. These prompts allow pre-trained models to adapt to new tasks with minimal retraining. In black-box scenarios, recent work~\cite{oh2023blackvip,ouali2023black,chilearning,chiadapting} eliminates the need for access to model weights by using pre-computed image and text features, greatly reducing computational costs. Our Handwritten Text Recognition (HTR) method is inspired by CLIP \cite{radford2021learning}, utilizing prompt tuning to adapt efficiently to individual handwriting styles, particularly in few-shot scenarios. Jia et al.\cite{jia2022visual} extended this concept to vision transformers through Visual Prompt Tuning (VPT), showcasing its effectiveness for visual recognition. Liu et al.~\cite{liu2024test} further extended this idea by using prompt padding as convolutional prompts for CNN-based models.


\noindent{\bf Meta-Learning}. 
Meta-learning has played an important role in many fields, including accelerating few-shot learning capabilities and adapting swiftly to new activities~\cite{liu2022few,gu2022improving}. Representative meta-learning methods include the Model-Agnostic Meta-Learning (MAML) algorithm \cite{finn2017modelagnostic} as well as its extensions (e.g. Reptile ~\cite{nichol2018reptile} and Meta-SGD ~\cite{li2017metasgd}, which further extends the approach). The practical application of meta-learning has shown its success in character recognition by effectively enabling the fast adaptation of the learning model to various writing styles and character sets~\cite{bartler2022mt3}. For instance, works in \cite{otsuzuki2021meta,qiu2021meta} have demonstrated the application of meta-learning to the character recognition task. Recent studies integrating meta-learning with prompt learning~\cite{wang2022learning,park2024prompt} on general computer vision tasks, differing from our focus on handwriting recognition.

\noindent{\bf Masked Autoencoders}. Masked Autoencoders (MAEs) are self-supervised neural networks designed to reconstruct missing parts of input data, such as segments of handwritten text \cite{he2022masked}. They work by encoding the visible portions into a latent representation and decoding it to predict the masked sections. Notable examples include BERT \cite{kenton2019bert}, which masks words for text prediction, and the Vision Transformer (ViT) \cite{yin2022vit}, which reconstructs masked image patches. MAEs are particularly useful in handwriting recognition, as they learn comprehensive representations from incomplete data, improving the understanding of diverse handwriting styles. Pengyuan Lyu’s work \cite{lyu2022maskocr} demonstrates that MAEs can effectively combine visual and linguistic information for more accurate text recognition. 

\noindent{\bf Test-time Adaptation}. Test-time adaptation (TTA) adjusts a pre-trained model to handle new, unseen data during inference, particularly when the data distribution differs from the training set \cite{chen2023improved,chi2021test,zhong2022meta,liu2023meta,wu2023metagcd,wu2024test,wang2024distribution}. TTA updates model parameters or internal statistics based on the test data without requiring ground truth labels. A notable TTA method is Tent (Test-time Entropy Minimization) \cite{wang2020tent}, which minimizes prediction uncertainty. Another approach involves self-supervised learning \cite{chen2022contrastive}, where the model fine-tunes itself using pseudo-labels generated from the test data. We employ TTA techniques to enhance model robustness and generalization under distribution shifts between training and test data \cite{yuan2023robust}. 

\section{Problem Definition}

\begin{figure*}
    \centering
    \includegraphics[width=1\linewidth]{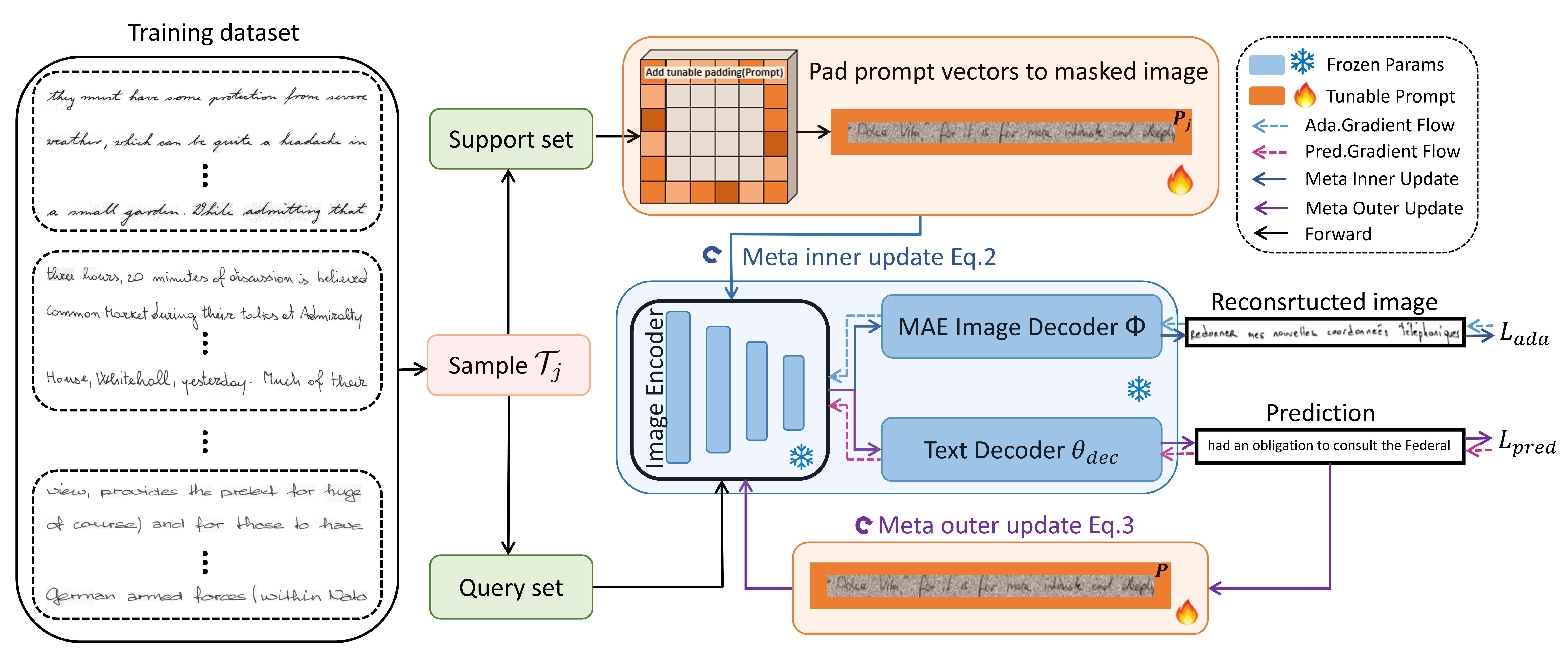}
    \vspace{-23pt}
    \caption{\textbf{Illustration of our approach during training.} The handwritten texts from a specific writer are divided into an unlabeled support set and a labeled query set. The images in the support set are masked, padded with meta prompt vectors, and passed through a shared image encoder, followed by reconstruction using a Masked Autoencoder (MAE)'s decoder. The writer-specific prompt vectors \( P_j \) are derived in the inner loop using a self-supervised loss \( \mathcal{L}_{\text{ada}} \), which optimizes the meta prompt vectors through a single gradient step. These writer-specific prompt vectors are then padded with the document images from the query set and used as input to the HTR model to predict a sequence of tokens representing the writing content. In the outer loop, the learned meta prompts \( P \) are updated based on a supervised loss \( \mathcal{L}_{\text{pred}} \)  computed by comparing the predicted text tokens against the ground truth.}
    \label{Architecture}
\end{figure*}

We assume that each writer exhibits a unique writing style with a distinct distribution of input handwritten image. Our goal is to adapt the HTR model to a new writer during testing with only a few unlabeled images of that writer. Our training set is comprised of \textit{M} writers $\{\mathcal{T}_j\}_{j=1}^M$. An individual writer $\mathcal{T}_j$ in the training set contains $N_j$ labeled data: $\mathcal{T}_j = \{(x_i, y_i)\}_{i=1}^{N_j}$, where $(x_i, y_i)$ denotes the $i$-th input handwritten image and the corresponding output digital text label of the writer $\mathcal{T}_j$. 
At test time, the model encounters an unseen writer $\mathcal{T}_{test}$ that exhibits a distribution shift from the training writers. The model aims to adapt to this writer using only \emph{a few unlabeled examples}. A key difference between our setting and MetaHTR~\citep{bhunia2021metahtr} is that MetaHTR replies on \emph{labeled data} for writer adaptation. In contrast, our problem setting does not require any labeled data during testing and is more realistic in real-world scenarios.


\section{Proposed Approach}
Conventional HTR models expect to generalize across diverse writing styles but lack writer-specific personalization. Test-time adaptation~\cite{sun2020test} provides a personalization framework using limited examples collected at test-time. However, achieving effective handwriting personalization without labeled test-time  data in a parameter-efficient way remains a significant challenge. In this work, we propose a efficient framework that aims to personalize HTR models for an individual writing style using a few unlabeled examples collected at test-time. Figure~\ref{Architecture} shows an overview of our proposed framework.

\subsection{Model Architecture}
Inspired by previous work~\cite{coquenet2023dan}, we use an encoder-decoder architecture $f(\cdot; \theta)$ parameterized by $\theta = ({\theta}_{enc}, {\theta}_{dec})$ as our backbone HTR model.
The image encoder $f(\cdot; {\theta}_{enc})$ aims to extract textual and layout representations from the input handwritten image. Specifically, we use a Fully Convolutional Network (FCN) as the image encoder that transforms a writer-specific input image \({\textbf{X}}\) into a two-dimensional feature map~\cite{coquenet2022end}, consisting of 18 convolutional layers followed by 12 depthwise separable convolutional layers, then add positional information to the 2D feature map and convert it into a 1D feature sequence~\cite{coquenet2023dan}. To condition on the extracted handwritten representation, we apply a stack of 8 transformer-based modules as the decoder $f(\cdot; {\theta}_{dec})$ \cite{vaswani2017attention}, and convert the 1D feature sequence produced by the encoder into a series of tokens~\cite{coquenet2023dan}. A start-of-transcription token is prepended to the feature tokens, and the decoder can autoregressively generate the predicted text tokens, i.e. \(\hat{\textbf{Y}}=(\hat{y}_0, \hat{y}_1, \ldots, \hat{y}_{t-1})\), until the end-of-transcription token \(\langle\text{eot}\rangle\) ~\cite{coquenet2023faster}.


\noindent\textbf{Prompt Tuning for Writing-style Personalization}. Visual prompt tuning~\cite{li2021prefix, zhou2022learning, lester2021power} has been shown to be an effective parameter-efficient fine-tuning framework. We incorporate its principles to personalize writer-specific handwritten text recognition in this work. Prompt tuning allows the model to adapt to downstream tasks by optimizing only learnable prompt vectors added to the input space, while keeping the rest of the model frozen \cite{li2021prefix}. We assume that writer-specific style knowledge can be effectively encoded within these prompt vectors. Specifically, following  \cite{liu2024test},  we use image padding as the prompts for handwriting personalization, serving as a part of input to the convolutional layers.


The padding techniques like zero padding or reflect padding are commonly used to control the output feature map size by adding zeros or reflected values around the input prior to convolution \cite{liu2024test}. By adjusting the padding values, we can modify the intermediate feature embeddings in convolutional layers. Therefore, using padding as prompt vectors offers a promising approach for personalized handwriting style estimation.


In our model, instead of fixed padding values, we employ tunable parameters that are updated during adaptation using unlabeled examples collected at test time. This allows the image encoder to embed each writer's unique style into the prompt vectors. As a result, this will create a personalized model. Notably, the prompts comprise less than 1\% of the model parameters, making it highly efficient for test-time personalization while preserving the model’s lightweight nature.

\begin{figure}
    \centering
    \centering
  \begin{tabular}{@{\hskip 2pt}c@{\hskip 2pt}c@{\hskip 1pt}c@{\hskip 1pt}}
    \textbf{Origional Image} & \textbf{Masked Image} & \textbf{Reconstructed Image}  \\
    \begin{subfigure}{0.16\textwidth}
      \includegraphics[width=\linewidth]{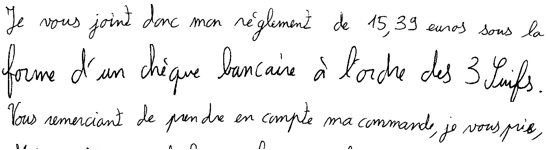}
      
    \end{subfigure} &
    \begin{subfigure}{0.16\textwidth}
      \includegraphics[width=\linewidth]{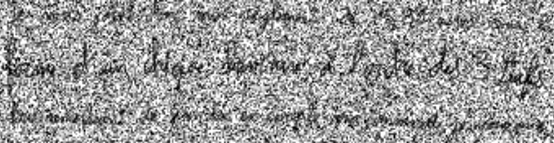}
     
    \end{subfigure} &
    \begin{subfigure}{0.17\textwidth}
      \includegraphics[width=\linewidth]{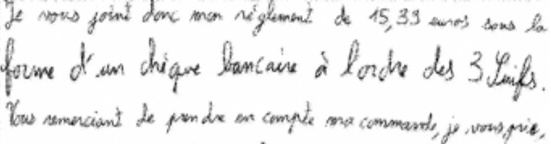}

    \end{subfigure} \\
    \begin{subfigure}{0.16\textwidth}
      \includegraphics[width=\linewidth]{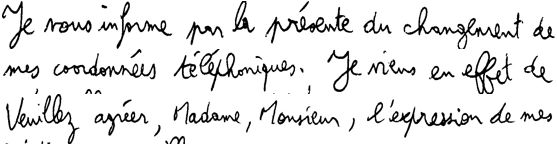}
      
    \end{subfigure} &
    \begin{subfigure}{0.16\textwidth}
      \includegraphics[width=\linewidth]{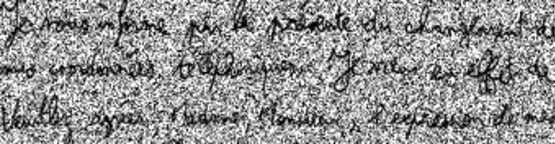}
      
    \end{subfigure} &
    \begin{subfigure}{0.17\textwidth}
      \includegraphics[width=\linewidth]{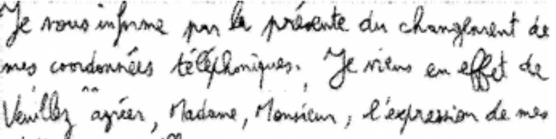}

    \end{subfigure} \\
    \begin{subfigure}{0.16\textwidth}
      \includegraphics[width=\linewidth]{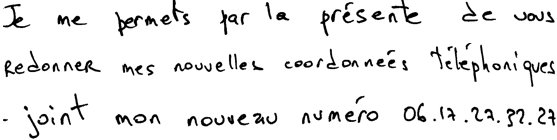}
      
    \end{subfigure} &
    \begin{subfigure}{0.16\textwidth}
      \includegraphics[width=\linewidth]{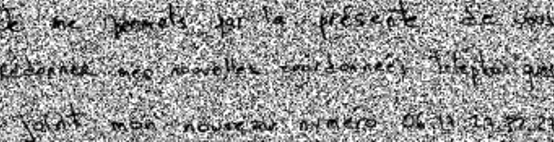}
       
    \end{subfigure} &
    \begin{subfigure}{0.17\textwidth}
      \includegraphics[width=\linewidth]{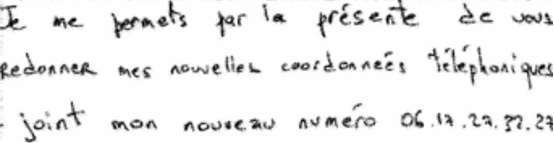}

    \end{subfigure} \\
    
  \end{tabular}
  \vspace{-8pt}
    \caption{\textbf{Illustration of the capability of Masked AutoEncoders to effectively tackle the HTR problem across three diverse handwriting styles from test data.} The first column represents the original input images. The second column displays the images masked by 75\%, which serve as the input for the reconstruction process. The third column presents the outputs, demonstrating how the MAE recovers the complete handwritten content from the masked images, effectively adapting to each unique writing style.
    }
    \label{style}
\end{figure}

\noindent\textbf{Test-time Adaptation with Unlabeled Examples}. To encourage prompt vectors to encode writer-specific style knowledge at test time, MetaHTR~\citep{bhunia2021metahtr} relies on few-shot labeled examples with supervised loss functions. To avoid the time-consuming annotation process, we instead introduce an auxiliary task with a self-supervised loss function. Specifically, we employ image reconstruction using a Masked Autoencoder (MAE) as an auxiliary task to personalize handwriting recognition. This auxiliary task aims to reconstruct an input handwritten image from a partially masked version with specific regions obscured. It encourages the model to infer missing content from the visible portions. This auxiliary task shares the backbone image encoder with the main HTR branch. We use $g(\cdot; \phi)$ to denote the MAE decoder parameterized by $\phi$.  An overview of MAE examples from three diverse handwriting styles is illustrated in Figure \ref{style}.


We employ the Structural Similarity Index Measure (SSIM) \cite{sara2019image}  as the self-supervised loss function for image reconstruction instead of Mean Squared Error (MSE), due to the high sensitivity of handwritten text recognition to image resizing, resolution changes, and minor details \cite{setiadi2021psnr}. SSIM is preferred as it assesses similarity based on texture and luminance, crucial for preserving perceptual quality in reconstructed images. Given a pair of original and reconstructed images \(x\) and \(\hat{x}\), the auxiliary loss \(L_{\text{ada}}\) is defined as:
\begin{equation}
L_{\text{ada}} = 1 - \text{SSIM}(x, \hat{x}) = 1 - \frac{(2\mu_x \mu_{\hat{x}} + a)(2\sigma_{x\hat{x}} + b)}{(\mu_x^2 + \mu_{\hat{x}}^2 + a)(\sigma_x^2 + \sigma_{\hat{x}}^2 + b)}
\label{eq:per0}
\end{equation}
where \(\mu_x\) and \(\mu_{\hat{x}}\) are the mean intensity values of images \(x\) and \(\hat{x}\), \(\sigma_x^2\) and \(\sigma_{\hat{x}}^2\) are their variances, and \(\sigma_{x{\hat{x}}}\) is their covariance. Constants \(a\) and \(b\) stabilize the calculation, ensuring robust performance. This SSIM-based loss better preserves the fine details of handwriting compared to MSE, enhancing the accuracy of text recognition.


At test-time, given a pre-trained image encoder $f(\cdot; {\theta}_{enc})$ and a MAE decoder $g(\cdot; {\phi})$, we freeze their parameters and only update the prompt vectors. We denote the prompt vectors as $ P \in \mathbb{R}^{ l \times 3}$, where $l$ is the number of padding around each handwritten image and the dimension matches the number of image channels. Let $\mathcal{T}_{test}$ represent an unseen writer during testing. We have access to a small number of $k$ unlabeled handwritten images $X = \{x_1, x_2, \ldots, x_k\}$ for adaptation to this unseen writer. To construct the auxiliary image reconstruction task, we generate masked handwritten images $X^m$. We then pad the prompt vectors \( P \) around these unlabeled and masked images, yielding \( \hat{X}^m \). The writer-specific prompts $P'$ are obtained through a few gradient updates by minimizing the loss  \(L_{\text{ada}}\) as shown in Eq. \ref{eq:per1}:

\vspace{-12pt}
\begin{equation}
\begin{aligned}
    P' &\leftarrow {P} - \lambda_1 \mathcal{L}_{\text{ada}}(g(f(\hat{X}^m, {\theta}_{enc}); \phi), X), \\
    &\quad \text{where} \quad \hat{X}^m  = PAD([P, X^m ]).
\end{aligned}
\label{eq:per1}
\end{equation}
\vspace{-12pt}

\subsection{Meta-Prompt for Writer Adaptive HTR}


At test time, the model first updates prompt vectors using a few unlabeled handwritten images with a self-supervised MAE loss (Eq.~\ref{eq:per0}) and then performs inference using the adapted prompt vectors on the remaining images. However, the prompt gradient with respect to the MAE loss may not align with minimizing text prediction error. Inspired by \cite{bhunia2021metahtr}, we apply a meta-learning algorithm to learn optimal prompts initialization across training writers, named meta-prompts, which enables prompt tuning to jointly minimize self-supervised and text prediction losses for any unseen test writers. Additionally, meta-learning naturally facilitates rapid adaptation in few-shot settings, making it practical for deployment in resource-constrained scenarios. Therefore, we treat the test-time adaptation for one particular writer style as one episode and simulate large-scale and diverse episodes with the training dataset.

\begin{algorithm}
\caption{Meta Prompt Training}
\label{alg:meta_prompt}
\begin{footnotesize} 
\begin{algorithmic}[1]
\Require Training set: $\mathcal{T}$; HTR model's pretrained image encoder and text decoder: $f(\cdot; {\theta}_{enc})$, $f(\cdot; {\theta}_{dec})$; MAE's pretrained image decoder: $g(\cdot; \phi)$; Trainable meta prompts: $P$
\Require Learning rates: $\lambda_1$ and $\lambda_2$
\State Initialize the prompt $P$ with Gaussian noise.
\While{not converged}
    \State Sample a batch of writers $\{\mathcal{T}_j\}_{j=1}^M \sim \mathcal{T}$ 
    
        \For{each writer $\mathcal{T}_j$}
            \State Sample support and query set $(\textbf{x}_\mathcal{S})$, ($\textbf{x}_\mathcal{Q}, \textbf{y}_\mathcal{Q}$) $\sim \mathcal{T}_j $ 
            \State Mask and pad support set $\hat{\textbf{x}}^m_\mathcal{S} = [P, \textbf{x}^m_\mathcal{S}]$
            \State Evaluate self-supervised loss: $\mathcal{L}_{\text{ada}}(\hat{\textbf{x}}^m_\mathcal{S}, \textbf{x}_\mathcal{S}; f_{{\theta}_{enc}}, g_{\phi})$
            \State Adapt prompts $P_j \leftarrow P - \lambda_1 \nabla_P \mathcal{L}_{\text{ada}}$
            \State Pad query set $\hat{\textbf{x}}_\mathcal{Q} = [P_j, \textbf{x}_\mathcal{Q}]$

            \State Evaluate text prediction $\mathcal{L}_{pred}(\hat{\textbf{x}}_\mathcal{Q},\textbf{y}_\mathcal{Q}; f_{{\theta}_{enc}}, f_{{\theta}_{dec}})$
            \EndFor
        
        \State Meta update $P \leftarrow P - \lambda_2 \nabla_P \sum_{\mathcal{T}_j} \mathcal{L}_{pred}$
        
    \EndWhile
\State Output meta-learned prompt $P$
\end{algorithmic}
\end{footnotesize} 
\end{algorithm}
\vspace{-5pt}
Algorithm \ref{alg:meta_prompt} outlines our training pipeline as a bi-level optimization process.  Specifically, the meta prompt vectors $P$ are initialized with Gaussian noise. For each iteration, we randomly sample a batch of episodes, with each episode corresponding to one writer $\mathcal{T}_j$. In each episode, we sample two non-overlapping sets: an unlabeled support set $(\textbf{x}_\mathcal{S})$ and a labeled query set ($\textbf{x}_\mathcal{Q}, \textbf{y}_\mathcal{Q}$) (L3-L5). We then mask the unlabeled support set to construct an auxiliary image reconstruction task. To obtain writer-specific prompts $P_j$, we update the meta prompt vectors $P$ with a single gradient step in the inner loop using a self-supervised loss $\mathcal{L}_{\text{ada}}$, controlled by a learning rate $\lambda_1$ (L7-L8). In addition, the writer-specific prompts are padded with handwritten images from the disjoint labeled query set and used as input to the pre-trained HTR model to generate text prediction tokens $\hat{\textbf{y}}_\mathcal{Q}$. We use cross-entropy as the supervised loss function $\mathcal{L}_{pred}$ to compare the predicted text tokens against the ground-truth $\textbf{y}_\mathcal{Q}$ (L10). The supervised loss function is shown as: 
\vspace{-5pt}
\begin{equation}
\begin{aligned}
    \mathcal{L}_{\text{pred}} &= \mathcal{L}_{CE}(f(f(\hat{\textbf{x}}_\mathcal{Q}; {\theta}_{enc}); {\theta}_{dec}), \textbf{y}_\mathcal{Q})\\
    &\quad \text{where} \quad \hat{\textbf{x}}_\mathcal{Q}  = PAD([P_j, \textbf{x}_\mathcal{Q}]).
\end{aligned}
\label{eq:per2}
\end{equation}

In the outer loop, we compute the total supervised loss over labeled query sets from all sampled episodes and optimize the meta-prompts using gradient-based methods, controlled by a learning rate $\lambda_2$. Through this meta prompt learning, we can obtain a learned initial prompt $P$ that can effectively adapt to a new writer with only a few unlabeled examples via the self-supervised update in Eq.~\ref{eq:per2}.

\subsection{ Implementation Details}\label{sec:details}
To speed-up the learning process, we implement a training strategy with several stages.

\noindent\textbf{Pre-training.}  In the pre-training phase, our model is first trained on synthetic printed lines~\cite{biswas2021docsynth} to acquire feature extraction capabilities using a line-level OCR model~\cite{goel2023handwritten}, and then applies a curriculum learning strategy that starts with 90\% synthetic data, gradually introducing more writer-batched real samples while reducing synthetic data to 20\% over time~\cite{rouhou2022transformer,coquenet2022end}. The pre-training process begins with synthetic data and gradually transitions to writer-based episode training, focusing on writer-specific data over time. 
The model processes mini-batches and employs a combination of preprocessing, an MAE auxiliary task, data augmentation, and curriculum dropout techniques~\cite{coquenet2023dan}.

\noindent\textbf{Meta-Prompt Training.} During the meta-prompt training phase, prompts are initialized using a Gaussian distribution with a mean of 0 and a variance of 1, we replace the padding in the first 18 convolutional layers of the encoder, and the encoder's other parameters are kept frozen.

\noindent\textbf{Personalization.} During the personalization phase, we use a few-shot approach, consistently selecting specific images for each individual rather than relying on random sampling. In this process, only the prompts are optimized, while all other parameters remain fixed.

 Our model is trained on a single NVIDIA RTX 4090 GPU. We use the Adam optimizer starting with a learning rate of $10^{-4}$. The same set of hyperparameters is used across all datasets. 

\section{Experiments}

In this section, we present the experimental setup and results. We provide an overview of the datasets and setup, compare our approach with state-of-the-art methods, and conduct an ablation study to analyze the impact of individual components on our model performance.

\subsection{Dataset and Setup}
We evaluate the performance of our method using two benchmark datasets: RIMES~\cite{grosicki2011icdar},  and the IAM Handwriting Database~\cite{marti2002iam}. Both datasets are widely used in the field of handwritten document recognition. They include word samples with annotated writer information, allowing the creation of writer-specific meta-batches for episodic training~\cite{bhunia2021metahtr}. 

The RIMES2011 dataset, derived from the RIMES project~\cite{grosicki2011icdar}, features over 12,000 pages of handwritten letters from more than 1,300 writers, offering diverse writing styles usually used in writer identification tasks~\cite{finn2017modelagnostic}. we use the line-level partition released by the ICDAR 2011 competition.

The IAM Handwriting Database contains 13,353 isolated and labeled text lines and 115,320 isolated and labeled words, offering a robust dataset of unconstrained handwritten English text~\cite{marti2002iam}. Following~\cite{bhunia2021metahtr}, we utilize the standard partitioning for training, validation, and testing as provided for the IAM dataset. MetaHTR operates at the word level and evaluates performance using the Word Recognition Accuracy (WRA) metric~\cite{bhunia2021metahtr}, while our approach uses line-level segmentation and evaluates performance with Character Error Rate (CER) and Word Error Rate (WER).

\begin{table}[ht]
\centering

\begin{tabular}{lcc}
\hline
\textbf{Method} & \textbf{CER}$\downarrow$ & \textbf{WER}$\downarrow$\\
\hline
\multicolumn{2}{l}{\textit{Instance-specific training methods}}\\
Easter2.0~\cite{chaudhary2022easter2} & 6.21\% & --  \\
SFR~\cite{wigington2018start} & 6.4\% & 23.2\%  \\
Decouple AN~\cite{wang2020decoupled} & 6.4\% & 19.6\%  \\
SPAN~\cite{coquenet2021span}  & 5.45 \% & 19.83\% \\
kang et al.~\cite{kang2022pay}  & 4.67 \% & -- \\

VAN~\cite{coquenet2022end} & 4.32\% & 16.24\%  \\

TrOCR~\cite{fujitake2024dtrocr} & 4.22\%& --\\
\hline
\multicolumn{3}{l}{\textit{Writer-specific training methods}}\\
Wang et al.~\cite{wang2022fast}& 5.3\% & 18.5\% \\

Ours  & \textbf{3.36}\% & \textbf{10.32}\%  \\
\hline
\end{tabular}
\vspace{-6pt}
\caption{\textbf{Evaluation of MetaWriter on the line level IAM \cite{marti2002iam} test set.} Our method achieves outstanding performance with a 5-shot adaptation setup, and outperforms other state-of-the-art methods. }
\label{tab:IAM}
\end{table}

\vspace{-12pt}
\begin{table}[ht]
\centering

\begin{tabular}{lcc}
\hline
\textbf{Method} & \textbf{CER}$\downarrow$ & \textbf{WER}$\downarrow$ \\
\hline
SPAN~\cite{coquenet2021span}  & 3.81 \% & 13.8\% \\
Puigcerver et al.~\cite{puigcerver2017multidimensional}*  & 3.3 \% & 12.8\% \\
Coquenet et al.~\cite{coquenet2022end} & 3.04\% & 8.32\%  \\
DAN~\cite{coquenet2023dan} & 2.63\% & 6.78\%   \\
\hline
Ours  & \textbf{2.19}\% & \textbf{6.63}\% \\
\hline
\end{tabular}
\vspace{5pt}
\caption{\textbf{Evaluation of MetaWriter using the line level Rimes ~\cite{grosicki2011icdar} test set.} We use a 5-shot adaptation setup and compare with state-of-the-art methods. *The work in \cite{puigcerver2017multidimensional} uses some postprocessing to further improve the result. For comparison, we report the result without this postprocessing for \cite{puigcerver2017multidimensional}.}
\label{tab:rimes}
\end{table}






\begin{table}[t]
\centering
\renewcommand{\arraystretch}{1.4} 
\resizebox{1\columnwidth}{!}{%
\begin{tabular}{l|c|c|c|c|c}
\hline
\textbf{Methods} & TrOCR~\cite{fujitake2024dtrocr} & SPAN~\cite{coquenet2021span} & DAN~\cite{coquenet2023dan} &  MetaHTR~\cite{bhunia2021metahtr} & Ours \\ \hline
\textbf{Params (M)} & 334 & 19.2 & 7.6  & 1.7 & \textbf{0.08} \\ \hline
\end{tabular}%
}
\vspace{-6pt}
\caption{\textbf{Comparison of trainable parameters across different methods.} 
Each column represents a method, while the rows indicate the method name and 
the total number of trainable parameters, measured in millions (M).}
\label{tab:param_comparison}
\end{table}

\vspace{-15pt}

\subsection{Comparison with the SOTA}
In this section, we compare our method with state-of-the-art approaches using two benchmark datasets: RIMES and IAM. We demonstrate not only the superior accuracy of our approach in terms of both Character Error Rate (CER) and Word Error Rate (WER), but also highlight its efficiency in handling large-scale handwriting recognition tasks. Through these comparisons, we provide a detailed analysis of the effectiveness and computational efficiency of our model in contrast to existing methods.

\noindent\textbf{Effectiveness of Our Method}. We evaluate the performance of our method using the IAM and RIMES test set and compare it with several state-of-the-art approaches in Table \ref{tab:IAM} and Table \ref{tab:rimes}. 

For the IAM dataset (Table \ref{tab:IAM}), MetaWriter achieves a significant reduction in both CER 3.36\% and WER 10.32\%, outperforming existing instance-specific and writer-specific training methods. It is important to note that, unlike MetaHTR, which uses 16 labeled samples for adaptation during testing on the world level, our evaluation is conducted with a few-shot adaptation using unlabeled data on the line level. Our model achieves a Word Recognition Accuracy (WRA) derived from the Word Error Rate of $89.7\%$ on IAM and $93.4\%$ on RIMES, compared to $89.2\%$ and $93.4\%$, respectively, reported by MetaHTR~\cite{bhunia2021metahtr}. However, a direct comparison between these results is challenging due to differences in experimental settings and dataset granularity. MetaWriter demonstrates superior accuracy, showcasing its capability to effectively adapt to individual writing styles and improve transcription results.

For the RIMES dataset (Table \ref{tab:rimes}), there are rarely writer-specific results available for comparison. As a result, we include comparisons with instance-specific methods to provide context. MetaWriter achieves a CER of 2.19\% and a WER of 6.63\%, outperforming other high-performing methods such as DAN~\cite{coquenet2023dan} and Coquenet et al.~\cite{coquenet2022end}. These results underscore MetaWriter's strong performance and consistent ability to enhance recognition accuracy, even when compared against instance-specific methods.

These results underscore the superior personalization capability of our MAE-based approach. By achieving lowest CER and WER compared to existing methods on both the IAM and RIMES datasets, our method not only reduces the word-level error rate by approximately 6\% on the IAM dataset but also sets new performance standards with an exceptionally low parameter count, providing a lightweight, effective solution for personalized handwritten text recognition.

\noindent\textbf{Parameter Efficiency}. As highlighted in our efficiency analysis, the low number of tunable parameters is a key strength of our method. To prove
this, we exhibit the number of tunable parameters for each
method of Table \ref{tab:param_comparison}. Specifically, our approach requires only 0.08M parameters to be trainable during personalization, a stark contrast to the 5.9M parameters in the decoder and 1.7M parameters in the encoder that remain untouched. This means that, while many existing methods demand the adaptation of all 7.6M parameters (encoder + decoder)~\cite{coquenet2023dan}, our approach optimizes just only 1\%—of the total model parameters. This not only leads to significantly faster adaptation but also drastically reduces the computational resources required, making our method highly practical for deployment in resource-constrained environments. The ability to achieve effective personalization by prompts with such a small number of parameters is a key advantage of our approach.


\subsection{Ablation Studies}
\vspace{-10pt}
\begin{table}[ht]
\centering

\resizebox{\columnwidth}{!}{%
\begin{tabular}{l|cc|cc}
\hline
\textbf{Method} & \multicolumn{2}{c|}{\textbf{IAM}} & \multicolumn{2}{c}{\textbf{RIMES}} \\
                 & \textbf{CER}$\downarrow$ & \textbf{WER}$\downarrow$ & \textbf{CER}$\downarrow$ & \textbf{WER}$\downarrow$ \\
\hline
Baseline              & 4.14\% & 12.03\% & 2.92\% & 8.35\% \\
MetaWriter w/o MAE             & 3.93\% & 11.06\% & 2.73\% & 7.24\% \\
MetaWriter w/o Meta             & 3.63\% & 10.75\% & 2.51\% & 7.08\% \\
MetaWriter w/o Prompt             & 3.41\% & 10.22\% & 2.21\% & 6.56\% \\
\hline
MetaWriter (Ours)              & \textbf{3.32}\% & \textbf{10.21}\% & \textbf{2.13}\% & \textbf{6.55}\% \\
\hline
\end{tabular}%
}
\vspace{-6pt}
\caption{\textbf{Ablation study of three components (MAE, Meta and Prompt) in our proposed method.} We evaluate the influence of CER and WER on the IAM and RIMES datasets.}
\label{tab: Ablation}

\end{table}

\vspace{-15pt}
\begin{table}[ht]
\centering

\resizebox{\columnwidth}{!}{%
\begin{tabular}{l|cc|cc|c}
\hline
\textbf{\# Conv} & \multicolumn{2}{c|}{\textbf{IAM}} & \multicolumn{2}{c|}{\textbf{RIMES}} & \textbf{\# Param} \\
                 & \textbf{CER}$\downarrow$ & \textbf{WER}$\downarrow$ & \textbf{CER}$\downarrow$ & \textbf{WER}$\downarrow$ & \\
\hline
L=0              & 3.51\% & 10.63\% & 2.42\% & 6.89\% & 0 \\
L=1              & 3.47\% & 10.55\% & 2.29\% & 6.73\% & 4.6K \\
L=5              & 3.44\% & 10.48\% & 2.24\% & 6.65\% & 23K \\
L=10             & 3.39\% & 10.39\% & 2.19\% & 6.62\% & 46K \\
L=15             & 3.35\% & 10.32\% & 2.15\% & 6.57\% & 69K \\\hline
L=18             & \textbf{3.31}\% & \textbf{10.2}\% & \textbf{2.11}\% & \textbf{6.52}\% & 82K \\
\hline
\end{tabular}%
}
\vspace{-6pt}
\caption{\textbf{The impact of adding prompts to different numbers of convolutional layers.} 
We evaluate performance by CER and WER on the IAM and RIMES datasets.}
\label{tab:prompts}

\end{table}

\vspace{-15pt}
\begin{table}[ht]
\centering
\scriptsize 
\resizebox{\columnwidth}{!}{%
\begin{tabular}{l|cc|cc}
\hline
\textbf{\# Samp} & \multicolumn{2}{c|}{\textbf{IAM}} & \multicolumn{2}{c}{\textbf{RIMES}} \\
                 & \textbf{CER}$\downarrow$ & \textbf{WER}$\downarrow$ & \textbf{CER}$\downarrow$ & \textbf{WER}$\downarrow$ \\
\hline
K=1              & 3.46\% & 10.53\% & 2.31\% & 6.84\% \\
K=3              & 3.41\% & 10.43\% & 2.25\% & 6.75\% \\
K=5              & 3.36\% & 10.32\% & 2.19\% & 6.63\% \\\hline
All Samples             & \textbf{3.32}\% & \textbf{10.21}\% & \textbf{2.13}\% & \textbf{6.55}\% \\
\hline
\end{tabular}%
}
\vspace{-6pt}
\caption{\textbf{Evaluation of the influence of CER and WER for IAM and RIMES datasets across different support set size for
personalization.} We observe that increasing the number of samples generally leads to a decrease in error rates.}
\label{tab:simplesize}

\end{table}

In this section, we perform various ablation studies to provide additional analysis of our proposed method.

\noindent\textbf{Impact of Different Components}. In this ablation study, we evaluate different components of the MetaWriter method to assess their influence on CER and WER across the IAM and RIMES datasets, as shown in Table \ref{tab: Ablation}. We consider the following variants of our method.

\noindent\emph{Baseline}:  The baseline model excludes all three additional methods, serving as a control to demonstrate the error rates without enhancements, underscoring the value of the added components.

\noindent\emph{MetaWriter w/o MAE}: This model omits the MAE component, and exhibits higher CER and WER compared to the other components, emphasizing the critical role of the MAE in enhancing performance. 

\noindent\emph{MetaWriter w/o Meta}: This method removing the meta-learning component led to further improvements in CER and WER.  This result emphasizes the importance of meta-learning in learning an optimal initialization of the prompts. Highlight the effectiveness of meta-learning aligns the unsupervised loss with word error, resulting in a more stable and consistent personalization process.

\noindent\emph{MetaWriter w/o Prompt}: This method without prompts showed a significant performance boost, achieving lower CER and WER across both datasets. However, without prompts, the model cannot benefit from faster adaptation, as it requires full parameter updates, which come at the cost of increased time and computational resources.

\noindent\emph{MetaWriter (our model)}: Our method incorporates all proposed components, and achieves the best CER and WER results. Importantly, prompt tuning offers 20 times fewer parameter adaptations than configurations without it. This advantage makes the full MetaWriter model highly suitable for real-time applications and deployment on edge devices, balancing performance with adaptation speed and efficiency.

\begin{figure}
    \centering
    \includegraphics[width=1\linewidth]{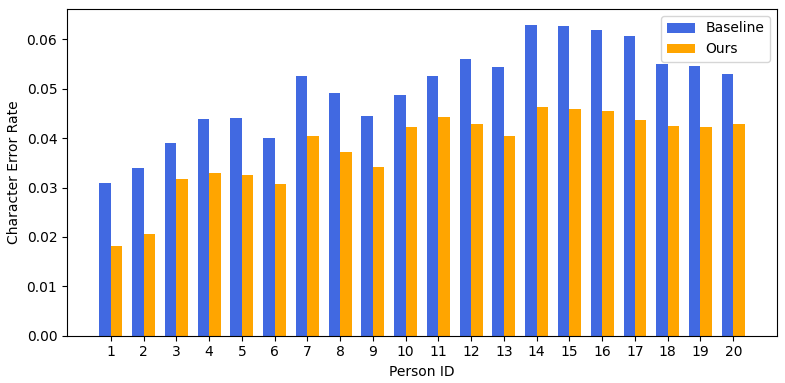}
    \vspace{-20pt}
    \caption{\textbf{illustration of the Character Error Rate (CER) for individual writers (Person IDs 1 to 20) on the IAM dataset.} We compare our method with the baseline, showing that our model improves accuracy across all writers.}
    \label{Person ID}
\end{figure}

\begin{figure}
    \centering
    \includegraphics[width=1\linewidth]{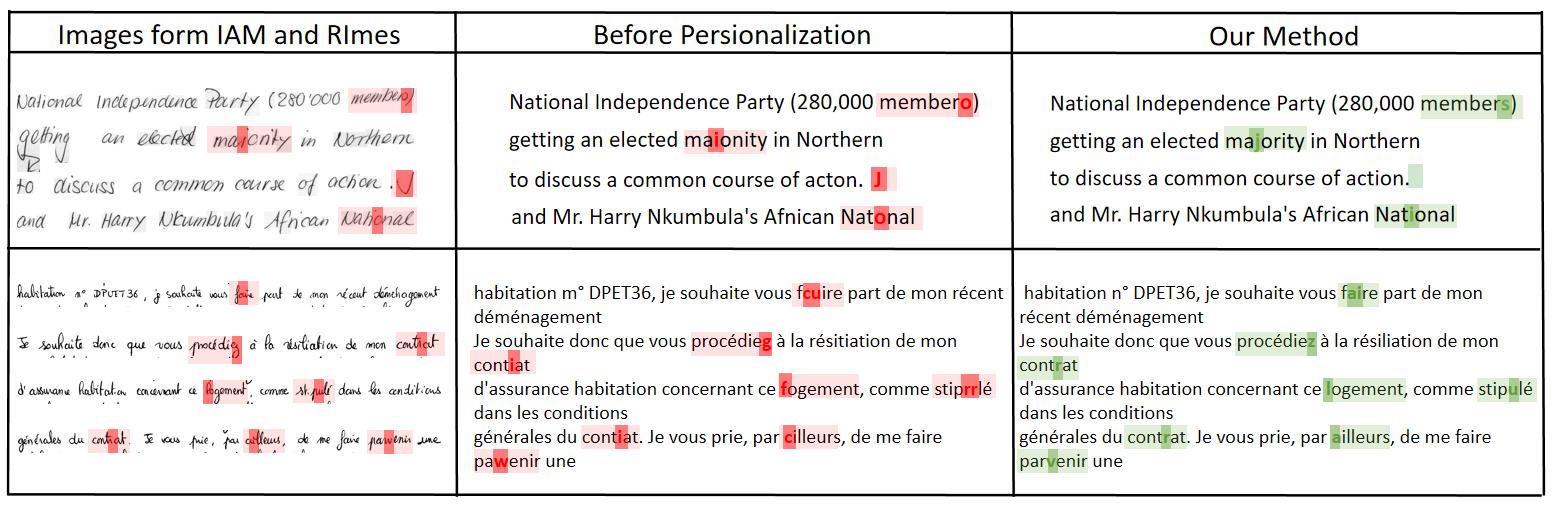}
    \vspace{-20pt}
    \caption{\textbf{Illustration of the handwriting transcription corrections by MetaWriter from IAM and RIMES.  
    }This figure shows original handwriting samples (left), the initial prediction marked with errors highlighted in red (center), and the corrected prediction after applying MetaWriter, highlighted in green (right). The first row represents samples from IAM, while the second row shows samples from RIMES. This demonstrates the effectiveness of our method in improving transcription accuracy.}
    \label{qualitative}
\end{figure}

\noindent\textbf{Impact of Prompt Layers}. To assess the impact of adding prompts to different numbers of convolutional layers, we evaluate configurations ranging from 0 to 18 layers, as shown in Table \ref{tab:prompts}. The results indicate that increasing the number of prompt layers led to improved CER and WER on the IAM and RIMES datasets, with the best performance at 18 layers. This improvement may be due to the structure of the encoder, which includes 18 convolutional layers followed by 12 depthwise separable convolutional layers. Prompts added to the first 18 layers influence the intermediate feature representation without directly affecting the final output, enabling effective adaptation. 

\noindent\textbf{Impact of Support Set Size}.
Table \ref{tab:simplesize} shows the impact of support set size on personalization performance for the IAM and RIMES datasets. CER and WER consistently decreased as the number of samples increased from 1 (K=1) to using all available samples. This trend indicates that using more samples enhances the model's ability to adapt effectively. Notable improvements were observed when increasing from one to five samples, with the best results achieved when all samples were used. This confirms that larger sample sizes contribute to better personalization and improved accuracy.

\noindent\textbf{Personalization Results}.
Figure~\ref{Person ID} shows CER for individual personalization on the IAM dataset, with each bar representing a person ID. Our method consistently achieves lower CERs than the baseline across all 20 individuals, highlighting its robustness in individual-specific adaptation. Figure~\ref{qualitative} further demonstrates our model's effectiveness in adapting to unique writing styles and maintaining high transcription accuracy across IAM and RIMES.

\noindent\textbf{Discussion}. Our work focuses on personalized HTR at the line level rather than the page level, as existing page-level datasets lack writer-specific metadata. Additionally, our techniques may not perform optimally at the page level, as simply padding with prompt vectors does not account for text location and reading order within a page.


\section{Conclusion}

In this paper, we propose a parameter-efficient framework for writer-specific personalization in Handwritten Text Recognition. Our approach formulates personalization as prompt tuning and integrates an auxiliary image reconstruction task with a self-supervised loss to facilitate adaptation using unlabeled test-time examples. To ensure that prompt tuning at test time minimizes both self-supervised and text recognition errors, we employ meta-learning to optimize prompt initialization. We demonstrate that our method outperforms state-of-the-art results on the RIMES and IAM benchmarks, achieving 20x fewer parameter adaptations and enabling practical deployment in resource-constrained settings.

{
    \small
    \bibliographystyle{ieeenat_fullname}
    \bibliography{main_plus}
}


\end{document}